\documentclass[journal=jacsat,manuscript=article]{achemso}

\usepackage{chemformula} % Formula subscripts using \ch{}
\usepackage[T1]{fontenc} % Use modern font encodings
\usepackage{placeins}
\usepackage{hyperref}
\usepackage[acronym,toc,shortcuts, nohypertypes={acronym}]{glossaries}
\newacronym[plural=MLPs]{MLP}{MLP}{Machine Learning Potential}
\newacronym{DFT}{DFT}{Density Functional Theory}
\newacronym[plural=RDFs]{RDF}{RDF}{Radial Distribution Function}
\newacronym[plural=ADFs]{ADF}{ADF}{Angular Distribution Function}
\newacronym[plural=PDOS]{PDOS}{PDOS}{phonon density of states}
\newacronym{MAE}{MAE}{mean absolute error}
\newacronym{RMSE}{RMSE}{root mean squared error}
\newacronym{GNN}{GNN}{graph neural network}
\author{Sebastien Röcken}
\author{Julija Zavadlav}
\email{julija.zavadlav@tum.de}
\affiliation{Professorship of Multiscale Modeling of Fluid Materials,
\linebreak
Department of Engineering Physics and Computation,
TUM School of Engineering and Design, 
Technical University of Munich, Germany}
\alsoaffiliation{Munich Data Science Institute $\&$ Munich Institute for Integrated Materials, Energy and Process Engineering, Technical University of Munich, Germany}
\email{julija.zavadlav@tum.de}

\title{Enhancing Machine Learning Potentials through Transfer Learning across Chemical Elements}

\begin{document}

\begin{tocentry}
\includegraphics[width=8.3cm]{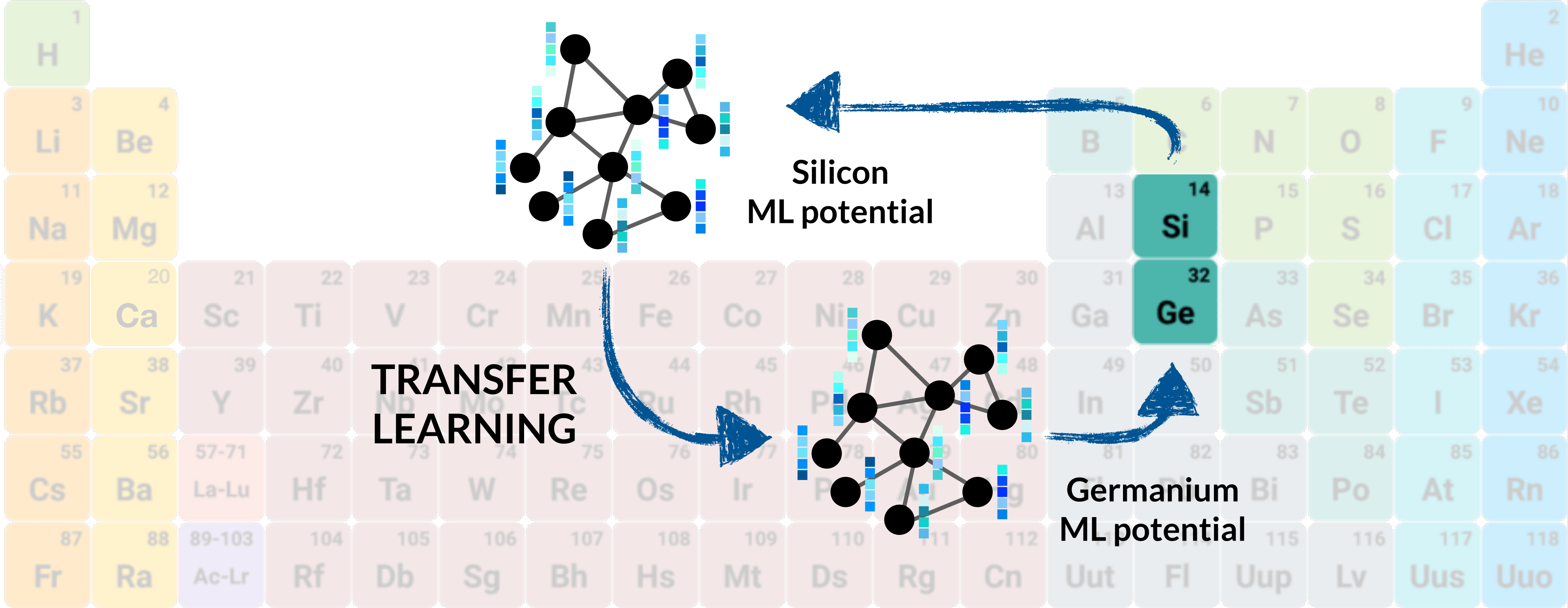}
%The surrounding frame is 9\,cm by 3.5\,cm, which is the maximum permitted for  \emph{Journal of the American Chemical Society} graphical table of content entries. The box will not resize if the content is too big: instead it will overflow the edge of the box.
\end{tocentry}

\begin{abstract}
Machine Learning Potentials (MLPs) can enable simulations of ab initio accuracy at orders of magnitude lower computational cost. However, their effectiveness hinges on the availability of considerable datasets to ensure robust generalization across chemical space and thermodynamic conditions. The generation of such datasets can be labor-intensive, highlighting the need for innovative methods to train MLPs in data-scarce scenarios. Here, we introduce transfer learning of potential energy surfaces between chemically similar elements. Specifically, we leverage the trained MLP for silicon to initialize and expedite the training of an MLP for germanium. Utilizing classical force field and ab initio datasets, we demonstrate that transfer learning surpasses traditional training from scratch in force prediction, leading to more stable simulations and improved temperature transferability. These advantages become even more pronounced as the training dataset size decreases. The out-of-target property analysis shows that transfer learning leads to beneficial but sometimes adversarial effects. Our findings demonstrate that transfer learning across chemical elements is a promising technique for developing accurate and numerically stable MLPs, particularly in a data-scarce regime.
\end{abstract}

\section*{INTRODUCTION}
Classical force fields are based on simple, physics-grounded functions, enabling rapid computation of atomic interactions~\cite{vanommeslaeghe2010charmm, wang2004development}. While they support large-scale Molecular Dynamics simulations, they often fall short in terms of accuracy. In contrast, simulations employing approximative solutions to the Schrödinger equation, such as \gls{DFT}, offer exceptional precision. Unfortunately, their computational demands render them infeasible to explore the extensive spatiotemporal scales typical of many complex systems.

\glspl{MLP} have emerged as a powerful tool to reconcile the trade-off between accuracy and computational efficiency~\cite{schutt2018schnet, batzner20223, musaelian2023learning, behler2007generalized, duval2023faenet, rocken2024accurate, rocken2024predicting}. The successful applications in literature span diverse systems ranging from alloys~\cite{bartok2018machine, wen2021specialising, rowe2018development, rocken2024accurate, andolina2023highly} to biological macromolecules~\cite{merchant2023scaling, takamoto2022towards}. Trained on \gls{DFT} data, \glspl{MLP} can achieve force prediction errors below \gls{DFT} accuracy~\cite{faber2017machine} while maintaining the ability to simulate million-atom systems~\cite{musaelian2023learning}. Despite these promising results, \glspl{MLP} are inherently limited by the scope of their training datasets and thus require large and informative datasets to yield reliable and accurate models. Various strategies have been proposed to address challenges related to data scarcity.

One widely used approach is active learning~\cite{smith2018less, thaler2024active, zhang2019active}, where new samples are iteratively selected, labeled, and incorporated into the training dataset. This process is typically driven by uncertainty quantification criteria, allowing models to identify which data points would be most beneficial for training. However, both data labeling and uncertainty quantification can be computationally intensive, especially when employing rigorous Bayesian methods~\cite{thaler2023scalable}. Although non-Bayesian uncertainty quantification techniques~\cite{kellner2024uncertainty, bigi2024prediction} may offer viable alternatives with lower computational costs, data labeling remains a notable hurdle in effective active learning implementation.

More recently, there has been a trend towards foundation \glspl{MLP}~\cite{takamoto2022teanet, takamoto2022towards, batatia2023foundation, allen2024learning,zhang2024dpa, shoghi2023molecules, yang2024mattersim, kolluru2022transfer}, where the application domain-unspecific models are built based on extensive datasets comprising a diverse array of chemical elements and compounds. These models can be fine-tuned on a specific chemical sub-space, ideally reducing the required training samples. However, this approach mandates a large-scale neural network to capture the interactions between elements of the entire periodic table, resulting in high computational costs for training and inference. As a result, \glspl{MLP} tailored to narrower chemical spaces or even focused on individual elements present a more computationally efficient alternative. Nevertheless, utilizing the transfer learning concept, the available databases covering different chemical space than the downstream application can still provide valuable information. 

In transfer learning, the knowledge gained from learning one task can be leveraged and transferred to a different but related task~\cite{zhuang2020comprehensive, pan2009}. If the transferred knowledge provides useful information, it can enhance data efficiency and accuracy. Therefore, this strategy is particularly advantageous when data for the target task is scarce. Transfer learning has been successfully applied across various domains, from image classification~\cite{oquab2014} to materials discovery~\cite{feng2021}. In the context of \glspl{MLP}, it was used to achieve higher accuracy by pre-training the model on the large DFT dataset and fine-tuning it on the more accurate but smaller coupled cluster dataset~\cite{dral2023learning, smith2019approaching, zaverkin2023transfer}. A similar idea is exploited by $\Delta$-Learning, where the difference between a less accurate and a more accurate model is learned~\cite{dral2023learning}.

Contrary to these previous studies, we propose to employ transfer learning across datasets that span different chemical spaces rather than fine-tuning the model on a chemical sub-space as in foundational models. Our objective is to leverage the knowledge gained while learning the potential energy surface of a chemical element to enhance the learning of another chemical element. Despite the differences in potential energy surfaces among different chemical elements, the fundamental interaction principles—such as steric and van der Waals forces—are common to all chemical elements. This shared foundation suggests that transfer learning of MLPs across chemical elements could be advantageous, particularly for chemical elements within the same group, as these elements exhibit similar chemical properties. This idea was recently briefly explored by Gardner et al.~\cite{gardner2024synthetic}. However, since they demonstrate the benefits of transfer learning only for force and energy predictions on synthetically altered data, transfer learning analysis is still missing for other properties and for unaltered real-case dataset scenario. 

In this paper, we examine transfer learning of \glspl{MLP} between chemical elements within the carbon group, specifically between silicon and germanium. Our investigation assesses the advantages of transfer learning across two distinct datasets spanning both solid and liquid phases. The first dataset utilizes the classical Stillinger-Weber potential, which allows for efficient data generation and enables us to rigorously examine the limits of transfer learning. The second dataset is a publicly available \gls{DFT} dataset, presenting a realistic application scenario. Through our analyses of the Stillinger-Weber dataset, we demonstrate that transfer learning significantly improves the accuracy of force predictions across the solid and liquid bulk regime, as well as enhances the accuracy of \gls{PDOS} in the small dataset regime. Notably, when training models using samples from a single temperature, we observe a marked enhancement in temperature transferability due to transfer learning. In the case of the \gls{DFT} dataset, transfer learning leads to better force prediction accuracy and more stable simulations compared to the models trained from scratch but does not appear to improve structural properties. Overall, our findings highlight the effectiveness of transfer learning across similar chemical elements as a valuable approach for developing \glspl{MLP}, especially as a tool to overcome the problems of sparse datasets.

%%%-------------------------------------------%%
\section*{METHODS}
In this section, we outline the methodology for training  \glspl{MLP} via force matching and describe the procedure for transfer learning among different chemical elements.

\subsection*{Machine Learning Potential Training via Force Matching}
The \gls{MLP} architecture utilized in this study is a message-passing \gls{GNN} DimeNet++~\cite{gasteiger2020fast}. This architecture incorporates directional information, which facilitates the inclusion of angular information, thereby enabling precise energy and force predictions. We employ our custom implementation as described in {\tt chemtrain}~\cite{fuchs2025chemtrain}. The embedding size is set to one-quarter of the original, and the cutoff distances are 0.5~nm for Stillinger-Weber data and 0.43~nm for \gls{DFT} data. The remaining structure adheres to the original implementation.

Different loss functions $L$ can be used to optimize the parameters $\theta$ of an \gls{MLP}~\cite{thaler2021learning,thaler2022}. Here, we use the force matching loss
\begin{equation}
    L (\theta) = \sum_{j=1}^{N_{bs}}\sum_{k=1}^{N} \sum_{l=1}^{3} \frac{1}{3N N_{bs}} (\hat{F}_{kl}(S_j) - F_{kl}(S_j;\theta))^2,
\end{equation}
where $N_{bs}$ is the batch size and $N$ is the number of atoms in a configurational state $S_j$ in the batch. $\hat{F}_{kl}$ and $F_{kl}$ denote the target and predicted force on atom $k$ in direction $l$, respectively. While it is more common to train also on energy labels, we do not use them here to avoid an additional hyperparameter and more complicated analysis. When training only on forces, the energy is determined only up to a constant. We evaluate this constant after the training by calculating the mean energy shift using the training data. The same constant is then used for energy predictions on the test dataset. As detailed in the Supplementary Information, we employ the Adam optimizer with learning rate decay for optimization. 

\subsection*{Transfer Learning}
\begin{figure}[h!]
    \centering
    \includegraphics[width=0.8\textwidth, trim= 0 7cm 24cm 0, clip ]{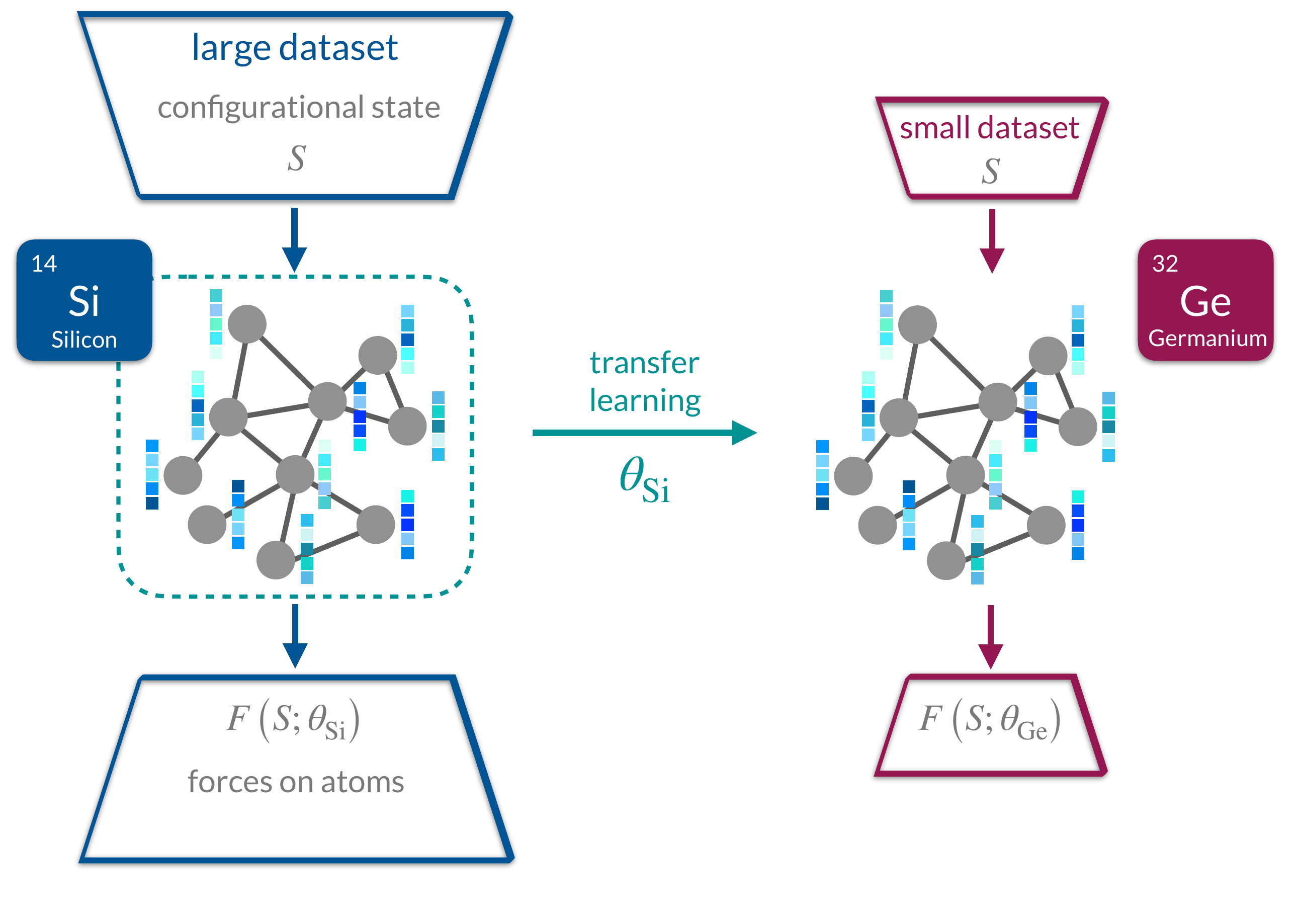}
    \caption{Transfer learning between chemical elements. An \gls{MLP} is initially trained on a large dataset containing various configurational states $\mathbf{S}$ and corresponding forces on atoms $\mathbf{F}$ for a certain chemical element, here silicon. The resulting parameters $\mathbf{\theta}$ are then transferred to initialize the parameters of an \gls{MLP}, which is subsequently trained on a small dataset of a similar but different chemical element, here germanium.}
    \label{fig:TL_procedure}
\end{figure}

The transfer learning of \glspl{MLP} across different chemical elements follows a two-stage procedure (Figure~\ref{fig:TL_procedure}). 
%%%-------------------------------------------%%
%%%-------------------------------------------%%
In the first stage, we pre-train the \gls{MLP} on a typically large dataset containing different structures and corresponding forces of a particular chemical element (e.g., silicon). In the second stage, we use the obtained parameters to initialize the fine-tuning of an \gls{MLP} for a similar chemical element (e.g., germanium), for which normally only a limited amount of data is available. We compare the performance of the transfer learning models with those trained from scratch that only undergo the second-stage training using random initialization.	

In the transfer learning process, it is possible to freeze some parameters. However, our preliminary analysis showed that the best accuracy is achieved when all parameters are fine-tuned. Therefore, during the second phase, we retrain all parameters of the \gls{MLP} for the new chemical element. The employed \gls{GNN} architecture encodes chemical elements via a learnable atom embedding vector. These embedding parameters are also transferred and retrained during the process.

\section*{RESULTS}
We evaluate the efficacy of transfer learning \glspl{MLP} across different chemical elements using Stillinger-Weber and \gls{DFT} datasets containing bulk silicon or germanium structures. The Stillinger-Weber data serves as a test example due to the simple potential energy surface and the ease of data generation, enabling accurate error estimation due to large test datasets. On the other hand, the \gls{DFT} data provides a realistic application scenario. In all cases, the direction of knowledge transfer is from silicon to germanium. 

%%%-------------------------------------------%%
\subsection*{SURROGATE MODEL FOR STILINGER-WEBER POTENTIAL}

\subsubsection*{Dataset}
We generate data for silicon and germanium with molecular dynamics simulations at 34 temperatures in the range 300 - 3600~K in 100~K intervals. The molecular interactions are parameterized using a Stillinger-Weber potential following Jian et al.~\cite{jian1990modification}. Simulations of a bulk system comprising 64 atoms are performed in LAMMPS~\cite{LAMMPS}, employing a time step of 1~fs, a Nose-Hoover thermostat with a damping factor of 100 time steps, and a Nose-Hoover barostat using a damping factor of 1000 time steps. For each temperature, we conduct a 4~ns NPT equilibration followed by a 1~ns NVT production simulation, sampling every 1~ps. We randomly sample these configurations to create different datasets. For silicon, the train, validation, and test dataset contain, respectively, 20, 5, and 10 samples per temperature or 680, 170, and 340 samples in total. For germanium, we discard the first 100~ps and create a data pool of 30.600 samples (900 per temperature) used for training, validation, and testing.

%%%-------------------------------------------%%
\subsubsection*{Data Efficiency}
To assess the benefits of alchemical transfer learning, we compare the accuracy of the transfer learning models with models trained from scratch for various training dataset sizes. The transfer learning models are first pre-trained on the silicon dataset. On the test dataset, the silicon \gls{MLP} model archives very low energy and force \gls{MAE} of 0.127~meV/atom and 0.69 meV/\AA, signaling an extensive training data size and a simple target molecular interaction. We then train (re-train in the case of transfer learning) several models on different numbers of samples from the germanium data pool to obtain the germanium \glspl{MLP}. 

\begin{figure}[h!]
    \centering
    \includegraphics[width=1\linewidth]{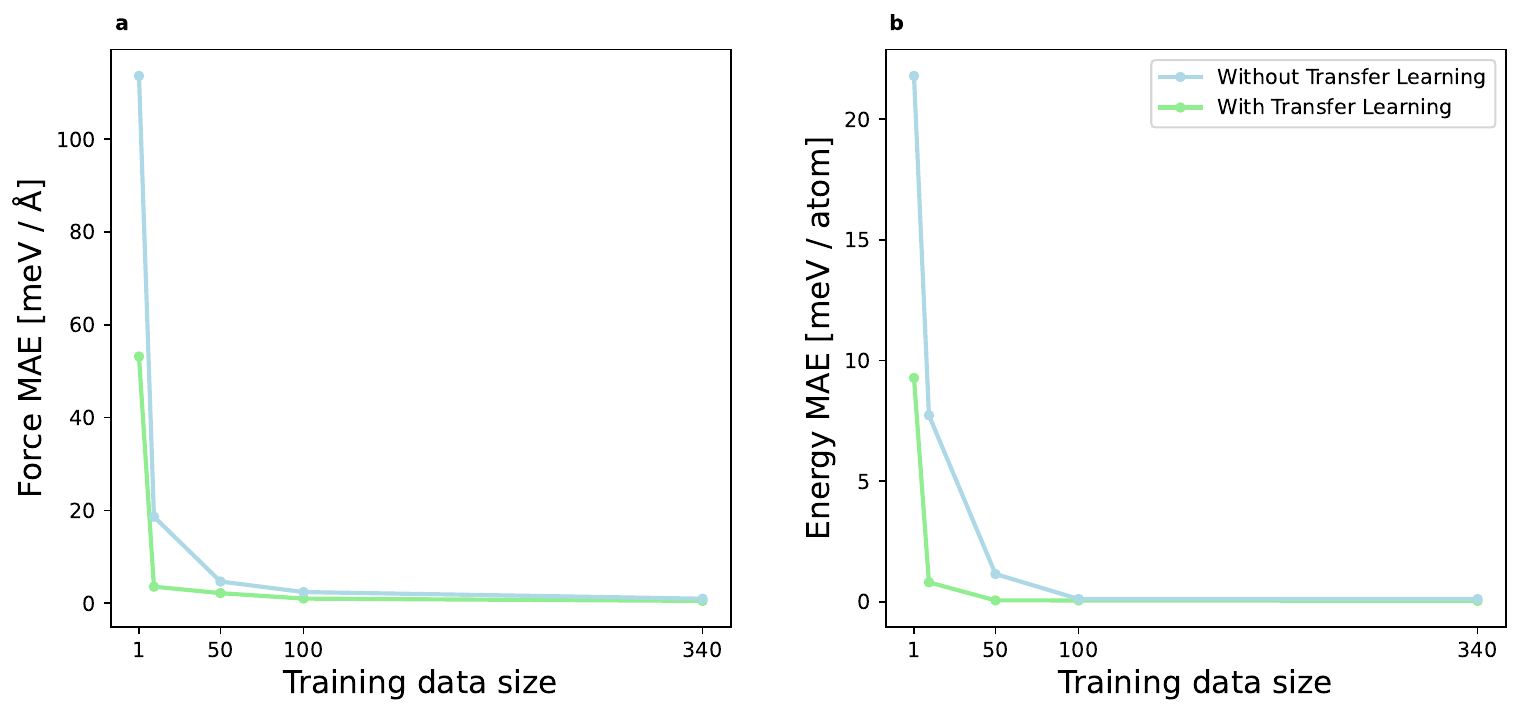}
    \caption{Data efficiency of transfer learning for Stillinger-Weber example. The green and blue lines denote the test set \gls{MAE} of force (a) and energy (b) predictions for the \glspl{MLP} trained with and without transfer learning, respectively. The \gls{MAE} values are averaged over five different models corresponding to different randomly selected train and validation data samples from the data pool. We perform a hyperparameter search for each mode based on the validation dataset (170 samples, 5 per temperature) as reported in the Supplementary Information. }
    \label{fig:trainsize_force_and_energy_300_to_3600K}
\end{figure}

In Fig.~\ref{fig:trainsize_force_and_energy_300_to_3600K}, we report the average \gls{MAE} of energy and force predictions evaluated on a fixed test dataset containing 340 configurations within the temperature range of 300-3600~K (10 samples per temperature). 
%%%
%%%
Transfer learning consistently outperforms training from scratch across all training sample sizes. We observe a significant positive transfer learning effect when germanium data is sparse. For example, transfer learning models trained with just 10 samples achieve a lower \gls{MAE} than models trained from scratch using 50 samples. We attribute this improvement to the similarity in the two underlying potentials and the broad distribution of silicon data. We repeated the analysis also for the solid state samples (see Supplementary Information) and found the same conclusions.

Next, we consider the performance for out-of-target properties and distributions. To achieve this, we further analyze the models trained with just one sample, as these models are already below the chemical accuracy threshold, both with and without transfer learning. Conversely, we will use the best model trained from scratch with the largest dataset of 340 samples as our reference model. At this training data size, the \glspl{MAE} have converged (Fig.~\ref{fig:trainsize_force_and_energy_300_to_3600K}), indicating that increasing the training dataset further would not significantly decrease the \gls{MAE}.

%%%-------------------------------------------%%
\subsubsection*{Material Properties}
As an out-of-target property, we evaluate the \gls{PDOS} for \gls{MLP} models trained on a single random snapshot at 2000~K. These configurations have an amorphous structure even though the temperature is above the germanium melting point due to the inaccuracies of the Stillinger-Weber potential and limited sampling. To compute the \gls{PDOS}, we generate a 2x2x2 germanium minimum energy super cell with 64 atoms in Avogadro~\cite{hanwell2012avogadro}. For code-specific reasons, we add normally distributed noise $\mathcal{N}(0, 10^{-6})$~nm to the particle positions to avoid exact 180° angles. \gls{PDOS} is then calcualted with phonopy~\cite{phonopy-phono3py-JPCM, phonopy-phono3py-JPSJ} using displacements of 0.01~Å.

\begin{figure}[h!]
    \centering
    \includegraphics[width=01.0\linewidth]{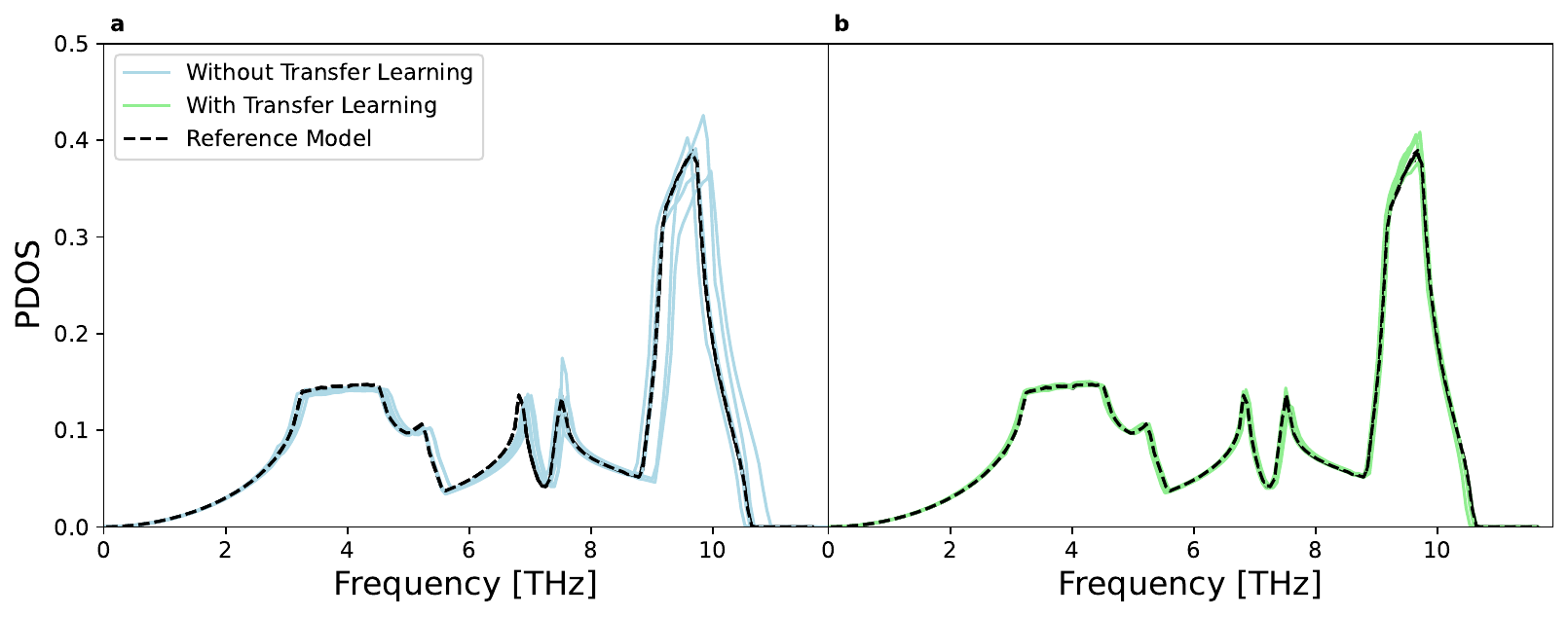}
    \caption{Phonon density of states (PDOS) for Stillinger-Weber example. The results are shown for five models obtained with (b, green) and without (a, blue) transfer learning. These models are trained with a single random sample at 2000~K and compared to the reference germanium model (dashed black) trained on a large dataset containing 340 samples across the entire considered temperature range. The five models correspond to the five best hyperparameter models with different random selections of training and validation data (5 samples at 2000~K).}
    \label{fig:pdos}
\end{figure}

The results illustrated in Fig.~\ref{fig:pdos}, demonstrate that models utilizing transfer learning closely match the reference \gls{MLP} that was trained from scratch on a large germanium dataset.
%%%
%%%
In contrast, models that do not use transfer learning show significant differences in \gls{PDOS} compared to the reference, with notable variations among the five models. This outcome is somewhat expected because the minimum energy configuration differs from the amorphous structure used during (re-)training. The enhanced accuracy observed with transfer learning suggests a successful knowledge transfer for the configurational distribution data gaps.

\subsubsection*{Temperature Transferability}
\begin{figure}[h!]
    \centering
    \includegraphics[width=0.9\linewidth]{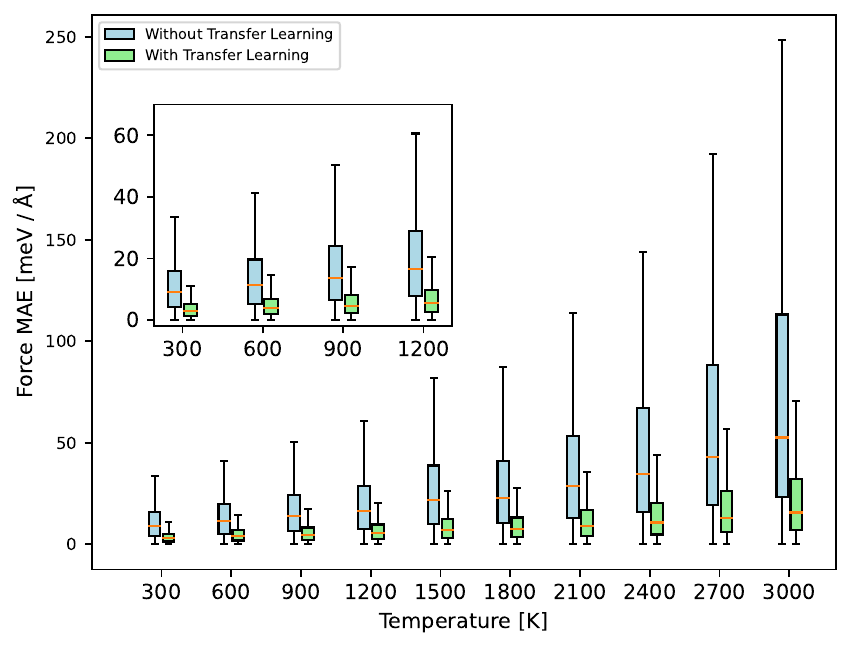}
    \caption{Temperature transferability for Stillinger-Weber example. Transfer learning models (blue) are referenced against the models trained from scratch (green) by computing the force \gls{MAE} for samples at different temperatures. For both cases, we train five models using a single sample at 2000~K and test each one on 50 samples at the respective temperature, resulting in 250 predictions in total for the 5 models. On the whisker plot, the orange line indicates the median, the box represents the interquartile range (IQR) between the first and third quartiles, and the whiskers extend to the furthest point within 1.5 times the IQR.}
    \label{fig:Temp_transferability}
\end{figure}

To further validate this point, we investigate the temperature transferability using the same models as above trained on a single sample at 2000~K. Figure~\ref{fig:Temp_transferability} displays the force \gls{MAE} per atom across test samples generated in the 300-3300~K temperature range. 
The transfer learning approach delivers significant gains across the entire temperature range. The silicon \gls{MLP} model initialization, parametrized with samples ranging from 300 to 3600 K, improves the germanium MLP's performance in temperature regimes not covered by the single germanium training sample. The same outcome is observed also when training on ten samples at 2000~K (see Supplementary Information).

\subsection*{DFT SURROGATE MODEL}
The results presented thus far were based on data generated using the Stillinger-Weber potential. While this example allowed us to rigorously evaluate performance with extensive test datasets, the Stillinger-Weber potential is relatively simple compared to the more accurate solutions provided by \gls{DFT}. Thus, we now proceed with a realistic use case by constructing a \gls{DFT} surrogate model. This model presents an ideal scenario for transfer learning, especially given the challenges associated with generating \gls{DFT} data.

\subsubsection*{Dataset}
We utilize the \gls{DFT} dataset for silicon and germanium provided by Zuo et al.~\cite{zuo2020performance}. Data includes 2x2x2 bulk structures with deformations up to ±20\%, single vacancies, and simulation data at temperatures ranging from 300~K to twice the melting temperature (2422~K for germanium and 3374~K for silicon). We exclude the slab structures from the dataset since these are not relevant to our analysis. To avoid exact 180° angles due to code-specific reasons, we add normally distributed noise $\mathcal{N}(0, 10^{-6})$~nm to the particle positions. We partition the silicon data into 182 training samples, 20 validation samples, and 23 test samples. For germanium, 24 samples are set apart for testing, while the remaining 217 samples are used as a data pool for training and validation.

\subsubsection*{Data Efficiency}
\begin{figure}[htb!]
    \centering
    \includegraphics[width=1\linewidth]{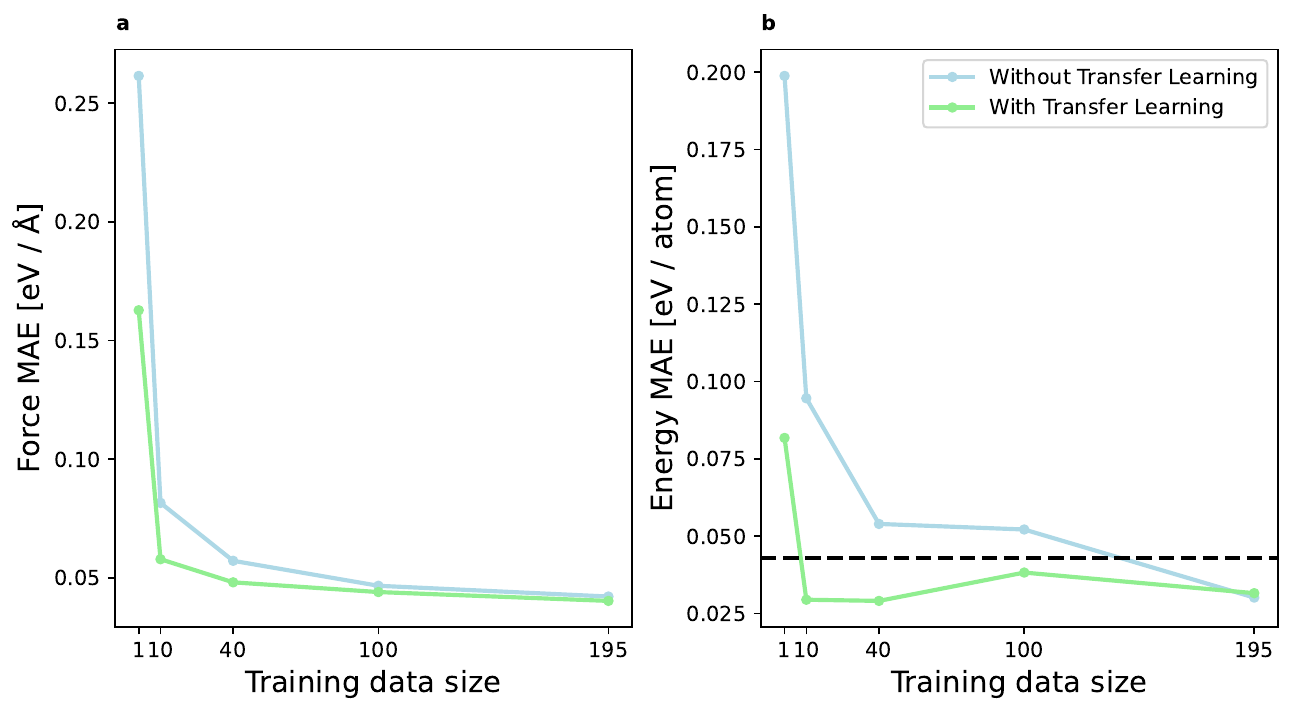}
    \caption{Data efficiency of transfer learning for the DFT example. The green and blue lines denote the test set \gls{MAE} of force (a) and energy (b) predictions for the \glspl{MLP} trained with and without transfer learning, respectively. The \gls{MAE} values are averaged over five different models corresponding to different randomly selected train and validation data samples. We perform a hyperparameter search for each model based on the validation dataset (22 samples) as reported in the Supplementary Information. The black dashed line denotes the chemical accuracy of 43~meV/atom.}
    \label{fig:Forces_and_Energies_DFT_TL_vs_no_TL_data_efficiency}
\end{figure}

We first test the accuracy of the silicon \gls{MLP} model. We obtain a similar test set accuracy as Zuo et al.~\cite{zuo2020performance}, confirming an expressive enough GNN and an adequate training strategy. In particular, the energy and force \gls{RMSE} are 21.8~meV/atom and 0.10~eV/\AA, respectively. 
Analogous to the Stillinger-Weber example, we proceed with the data efficiency testing of the transfer learning approach by reporting for germanium MLP models the force and energy \gls{MAE}  using 1, 10, 40, 100, and 195 training samples (Fig.~\ref{fig:Forces_and_Energies_DFT_TL_vs_no_TL_data_efficiency}). 
Again, we find a significant performance enhancement when employing the transfer learning approach. For the respective training sample sizes, the energy \gls{MAE} is reduced by 59, 69, 46, 27, -5\% and the force \gls{MAE} by 38, 29, 16, 6, 4\%, highlighting positive transfer learning effects for scarce data. With only 10 training samples, the transfer learning models achieve an accuracy below the chemical threshold of 43~meV. In contrast, training from scratch requires all available data (195 samples) to reach chemical accuracy. Note that at this training data size, the \gls{MAE} of the models with and without transfer learning is very similar and comparable to the errors reported by Zuo et al.~\cite{zuo2020performance}

\subsubsection*{Structural Properties}
\begin{figure}[h!]
    \centering
    \includegraphics[width=1.0\linewidth]{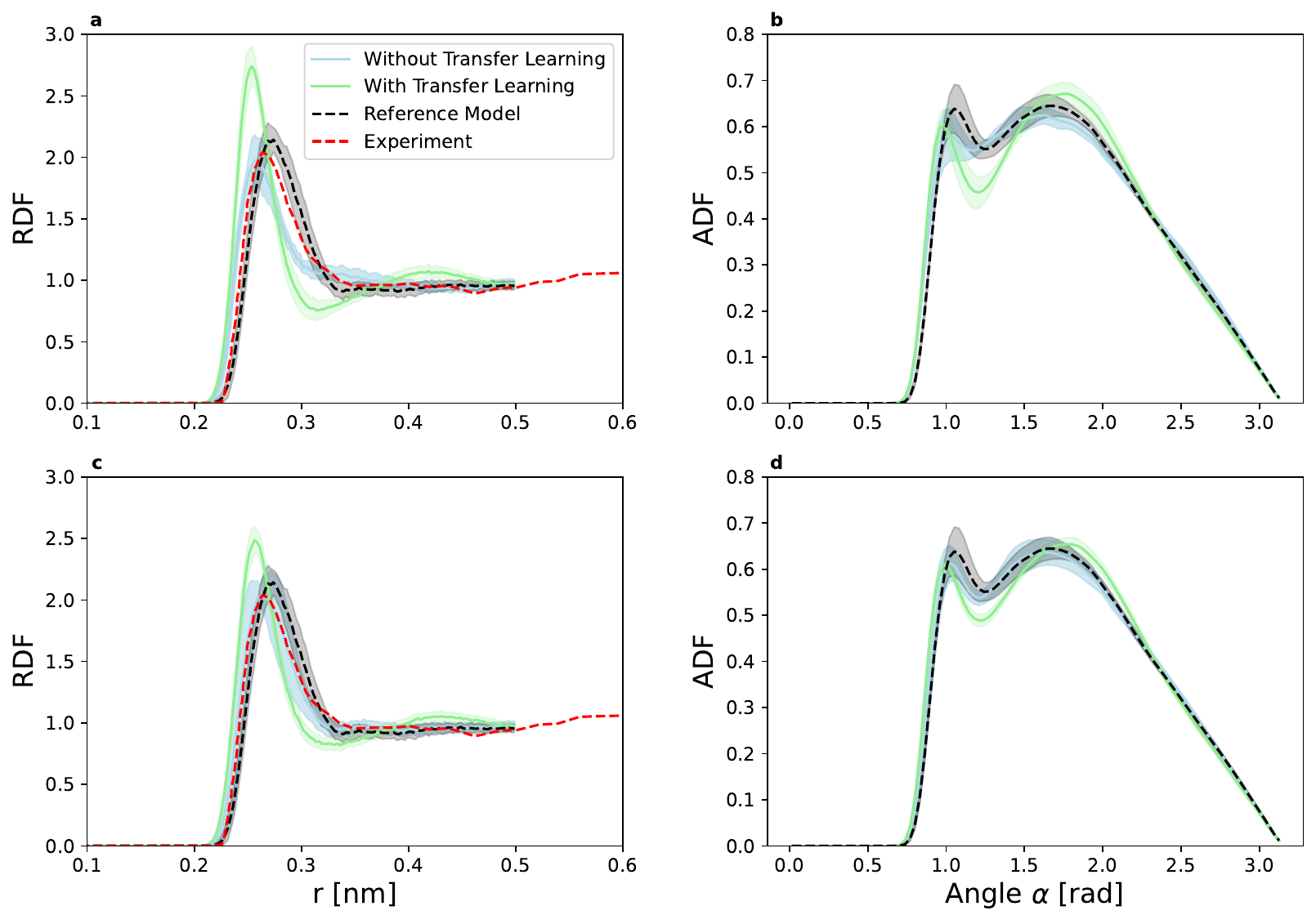}
    \caption{Radial Distribution Function (RDF; left) and Angular Distribution Function (ADF; right) for the DFT example and training data size of 10 (a, b) and 100 (c, d). The green and blue lines denote the models employing transfer learning and the models trained from scratch, respectively. As a reference, we show the results also for the reference model (dashed black), i.e., a germanium MLP trained from scratch with all available data (195 samples). Since some simulation runs resulted in unphysical trajectories, we computed the \gls{RDF} and \gls{ADF}  from states sampled in the time regime of 10-100~ps, only for models that ran for the full 100~ps and yielded zero \gls{RDF} values below 0.2~nm. The black solid line represents experimental results from neutron diffraction studies at 1273.15~K~\cite{salmon1988neutron}. Shaded regions denote the standard deviation.}
    \label{fig:DFT_sim_TL_vs_no_TL_10_100_195_samples}
\end{figure}

We further assess the impact of transfer learning by examining the structural properties of liquid germanium. In particular, the \glspl{RDF} and \glspl{ADF}, shown in Fig.~\ref{fig:DFT_sim_TL_vs_no_TL_10_100_195_samples}. 
%%%
%%%
The structural properties are evaluated from 100 forward simulations (20 velocity initializations for each of the five models differing in the train/validation data split). The simulations are performed under NVT conditions for 100~ps using the Velocity Verlet integration and a time step of 0.5~fs. The 1200~K temperature is maintained with the Langevin thermostat with a damping coefficient of 1~/ps. All simulations start with the same liquid state configuration in the dataset containing 64 atoms. 

With scarce data, the structural properties are not yet fully converged for both approaches—those that use transfer learning and those that do not. Additionally, the \glspl{RDF} show some deviation from the experimental curve, even with the largest training dataset. This suggests either inaccuracies in \gls{DFT} calculations or insufficient data. Interestingly, models trained from scratch display a closer alignment with the experimental data, revealing a negative transfer learning effect for the structural properties, despite having lower errors in energy and force predictions. For models that utilize transfer learning, the \glspl{RDF} and \glspl{ADF} gradually shift from the values obtained using the pre-trained silicon model to those derived from the reference germanium model trained from scratch (see Supplementary Information Fig.~3). This finding effectively illustrates the mechanics of transfer learning: the models revert to the pre-trained solution in regions where training data is limited. Although the pre-trained silicon model differs from the target germanium model, it is physically valid and contributes to improved numerical stability, as demonstrated in the following section.

\subsubsection*{Numerical Stability}
Over the past decade, the development of \glspl{MLP} has advanced significantly. However, numerical stability remains a crucial area that requires improvement~\cite{fu2022forces}. Specifically, \glspl{MLP} can display pathological behavior, such as extreme energy and force predictions, resulting in unphysical trajectories characterized by unrealistic bond breaking, particle overlaps, and unstable simulations that may lead to simulation failure.

The stability of MLPs is typically assessed using structural criteria, such as measuring deviations from equilibrium \glspl{RDF} or equilibrium bond lengths~\cite{fu2022forces}. In our analysis, we examined 100 forward trajectories used for \gls{RDF} and \gls{ADF} computation and evaluated the number of simulations that successfully completed 100~ps without any RDF values falling below 0.2~nm, which signifies no particle overlaps (see Table~\ref{Tab:successful_simulations_DFT}).
%%%
% Tab successful simulations
\begin{table}[h!]
    \centering
    \begin{tabular}{c|c|c|c|c}
         & \multicolumn{4}{|c}{Training data size} \\
        \cline{2-5}
        Approach & 10 & 40 & 100 & 195 \\
        \hline
        Without Transfer Learning & $35$ & $46$ & $73$ & $71$ \\
        With Transfer Learning & $100$ & $100$ & $80$ & $80$ \\
    \end{tabular}
    \caption{Numerical Stability of DFT surrogate models with and without transfer learning and various training data sizes. The reported values correspond to the number of successful simulations out of 100 that reached 100~ps without exploding and displaying a zero \gls{RDF} value below 0.2~nm.}
    \label{Tab:successful_simulations_DFT}
\end{table}
%%%

We find that transfer learning significantly improves the numerical stability of \glspl{MLP}, particularly when working with small training datasets. When conducting forward simulations of liquid germanium using MLPs trained on limited samples, we are likely to encounter out-of-distribution states. However, by employing transfer learning from a silicon \gls{MLP} with a broader training data distribution, we can access approximate and physically sound information in these unseen regimes, resulting in more stable behavior. This finding supports our overarching hypothesis that pre-training the model on a chemically similar system allows for the transfer and preservation of information from the pre-training dataset, which is absent from the fine-tuning dataset. The positive impacts of transfer learning are thus most pronounced when there is a significant difference between the distributions of the pre-training and fine-tuning datasets.

\FloatBarrier
%%%-------------------------------------------%%
\section*{CONCLUSIONS}
In conclusion, our study highlighted the advantages of the transfer learning technique in the development of \glspl{MLP}. We built upon the idea that the underlying physical principles governing the interactions of atoms are shared. The more similar two chemical elements are, the closer their potential energy surfaces are likely to be. This reasoning underpins the assumption that transfer learning is beneficial when training \glspl{MLP} between similar chemical elements. Indeed, we found many benefits of transfer learning \glspl{MLP} from silicon to germanium, which are closely related and share the same crystal structure. It enables a significantly higher accuracy in force and energy predictions in scenarios with limited training data, supporting the hypothesis that transfer learning among similar chemical elements enhances data efficiency and accuracy. Additionally, we present examples of both positive and negative transfer learning outcomes for other property predictions, allowing for a deeper understanding of the transfer learning mechanism. Importantly, we demonstrate that transfer learning facilitates more stable simulations, suggesting that this approach could play a key role in overcoming numerical stability challenges associated with MLPs.

\subsection*{Data Availability}
The data that supports the findings of this study are available within the article and its supplementary material. The code to train presented models will be open-sourced at GitHub: \href{https://github.com/tummfm/TL\_MLP.git}{https://github.com/tummfm/TL\_MLP.git}.

\begin{acknowledgement}
This research was funded by the Deutsche Forschungsgemeinschaft (DFG, German Research Foundation) - 534045056. The authors thank Sunny Tamrakar and Frederico Pita de Araujo for their contributions to the initial feasibility studies.
\end{acknowledgement}

\begin{suppinfo}

The Supporting Information is available free of charge. Hyperparameter for training on Stillinger-Weber data (Tab. S1); Hyperparameter for training on DFT data (Tab. S2); Data efficiency on test samples in the 300-900~K range for Stillinger-Weber data (Fig. S1); Temperature transferability when training on 10 samples for Stillinger-Weber data (Fig. S2); Structural properties for all transfer learning models on the \gls{DFT} dataset (Fig. S3).
\end{suppinfo}

\bibliography{references}

\end{document}